\documentclass{article} 
\usepackage{iclr2023_conference,times}

\usepackage{amsmath,amsfonts,bm}

\def\eqref#1{equation~\ref{#1}}

\def\1{\bm{1}}

\DeclareMathAlphabet{\mathsfit}{\encodingdefault}{\sfdefault}{m}{sl}
\SetMathAlphabet{\mathsfit}{bold}{\encodingdefault}{\sfdefault}{bx}{n}

\usepackage{hyperref}
\usepackage{url}
\usepackage{graphicx}
\graphicspath{ {./figures/} }
\usepackage{shortbold}

\newcommand{\cut}[1]{}

\usepackage{verbatim}
\usepackage[inline]{enumitem}
\usepackage{graphbox}
\usepackage{listings}
\usepackage{algorithm}
\usepackage{algpseudocode}

\title{Contrastive Meta-Learning for \\ Partially Observable Few-Shot Learning}

\author{Adam Jelley\textsuperscript{1}, \quad Amos Storkey\textsuperscript{1}, \quad Antreas Antoniou\textsuperscript{1}, \quad Sam Devlin\textsuperscript{2}\\
\textsuperscript{1}School of Informatics, University of Edinburgh, \quad \textsuperscript{2} Microsoft Research Cambridge}

\iclrfinalcopy 
\begin{document}

\maketitle

\begin{abstract}
Many contrastive and meta-learning approaches learn representations by identifying common features in multiple views. However, the formalism for these approaches generally assumes features to be shared across views to be captured coherently. We consider the problem of learning a \textit{unified representation from partial observations}, where useful features may be present in only some of the views. We approach this through a probabilistic formalism enabling views to map to representations with different levels of uncertainty in different components; these views can then be integrated with one another through marginalisation over that uncertainty. Our approach, \textit{Partial Observation Experts Modelling} (POEM), then enables us to meta-learn consistent representations from partial observations. We evaluate our approach on an adaptation of a comprehensive few-shot learning benchmark, Meta-Dataset, and demonstrate the benefits of POEM over other meta-learning methods at representation learning from partial observations. We further demonstrate the utility of POEM by meta-learning to represent an environment from partial views observed by an agent exploring the environment.\footnote{Implementation code is available at \url{https://github.com/AdamJelley/POEM}}
\end{abstract}

\begin{figure}[h!]
\centering
  \includegraphics[width=\textwidth]{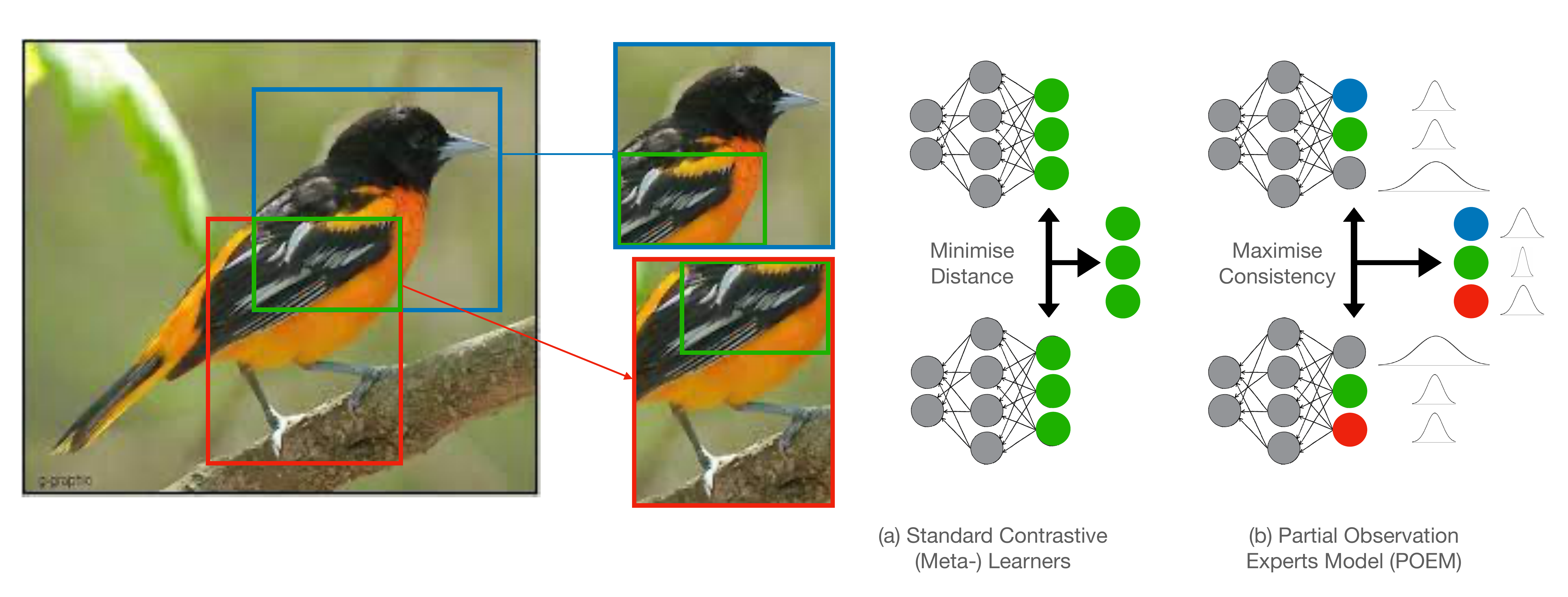}
  \vspace{-6mm}
  
  \caption[Figure 1 caption]%
  {Standard contrastive (meta-) learners minimise a relative distance between representations. This encourages the learning of features that are consistent in all views; in the above example this corresponds to the pattern on the bird's wing. To better handle partial observability, where features may be disjoint between views, we propose Partial Observation Experts Modelling (POEM). POEM instead maximises consistency between multiple views, by utilising representation uncertainty to learn which features of the entity are captured by a view, and then combining these representations together by weighting features by their uncertainty via a product of experts model \citep{hinton2002training}.}
\label{fig:POEM1}
\end{figure}

\section{Introduction} \vspace{-3mm}

Modern contrastive learning methods \citep{radford_learning_2021, chen_simple_2020, he_momentum_2020, oord_representation_2019}, and embedding-based meta-learning methods such as Prototypical Networks \citep{snell_prototypical_2017, vinyals_matching_2016, sung_learning_2018, edwards_towards_2017}, learn representations by minimizing a relative distance between representations of related items compared with unrelated items \citep{ericsson_self-supervised_2021}. However, we argue that these approaches may learn to disregard potentially relevant features from views that only inform part of the representation in order to achieve better representational consistency, as demonstrated in Figure~\ref{fig:POEM1}. We refer to such partially informative views as \textit{partial observations}. The difficulty with partial observations occurs because distances computed between representations must include contributions from all parts of the representation vector. If the views provided are diverse, and therefore contain partially disjoint features, their representations may appear different to a naive distance metric. For example, two puzzle pieces may contain different information about the whole picture. We call this the problem of \emph{integrative representation learning}, where we wish to obtain a representation that integrates different but overlapping information from each element of a set.

In this paper, we provide a probabilistic formalism for a few-shot objective that is able to learn to capture representations in partially observable settings. It does so by building on a product of experts \citep{hinton2002training} to utilise representation uncertainty: a high variance in a representation component indicates that the given view of the data poorly informs the given component, while low variance indicates it informs it well. Given multiple views of the data, the product of experts component in POEM  combines the representations, weighting by the variance, to get a maximally informative and consistent representation from the views.

To comprehensively evaluate our approach, we adapt a large-scale few-shot learning benchmark, Meta-Dataset \citep{triantafillou_meta-dataset_2020}, to evaluate representation learning from partial observations. We demonstrate that our approach, \emph{Partial Observation Experts Modelling} (POEM), is able to outperform standard few-shot baselines on our adapted benchmark, Partially Observed Meta-Dataset (PO-Meta-Dataset), while still matching state-of-the-art on the standard benchmark. Finally, we demonstrate the potential for our approach to be applied to meta-learn representations of environments from the partial views observed by an agent exploring that environment.

The main contributions of this work are: 
\begin{enumerate*}
\item A probabilistic formalism, POEM, that enables representation learning under partial observability; 
\item Comprehensive experimental evaluation of POEM on an adaptation of Meta-Dataset designed to evaluate representation learning under partial observability, demonstrating that this approach outperforms standard baselines in this setting while still matching state-of-the-art on the standard fully observed benchmark;
\item A demonstration of a potential application of POEM to meta-learn representations of environments from partial observations.
\end{enumerate*}

\section{Related Work}

\subsection{Contrastive Learning} 
Contrastive learning extracts features that are present in multiple views of a data item, by encouraging representations of related views to be close in an embedding space \citep{ericsson_self-supervised_2021}. In computer vision and natural language applications these views typically consist of different augmentations of data items, which are carefully crafted to preserve semantic features, and thereby act as an inductive bias to encourage the contrastive learner to retain these consistent features \citep{le-khac_contrastive_2020}. A challenge in this approach is to prevent representational `collapse', where all views are mapped to the same representation. Standard contrastive approaches such as Contrastive Predictive Coding \citep{oord_representation_2019}, MoCo \citep{he_momentum_2020}, and SimCLR \citep{chen_simple_2020} handle this by computing feature space distance measures relative to the distances for \emph{negative} views -- pairs of views that are encouraged to be distinct in the embedding space. In this work we take a similar approach, where the negative views are partial observations of distinct items, but we aim to learn to unify features from differing views, not just retain the consistent features. We learn to learn a contrastive representation from partial views. We note that state-of-the-art representation learning approaches such as CLIP \citep{radford_learning_2021}, which leverage contrastive learning across modalities, also suffer from extracting only a limited subset of features \citep{furst_cloob_2022} due to using an embedding-based approach \citep{vinyals_matching_2016} to match image and text representations.

\subsection{Embedding-Based Meta-Learning}
Embedding-based meta-learners similarly learn representations of classes by extracting features that are consistently present in the data samples (generally referred to as \emph{shots} in the meta-learning literature) provided for each class, such that the class of new samples can be identified with a similarity measure \citep{hospedales_meta-learning_2020}. These methods generally differ in terms of their approach to combine features, and the distance metric used. Prototypical Networks \citep{snell_prototypical_2017} use a Euclidian distance between the query representation and the average over the support representations for a class (referred to as a \textit{prototype}). Relation Networks \citep{sung_learning_2018} use the same prototype representation as Prototypical Networks, but use a parameterised \emph{relation module} to learn to compute the similarity between the query and the prototype rather than using a Euclidian distance. Matching Networks \citep{vinyals_matching_2016} use a Cosine distance between the query sample and each support sample as a weighting over the support labels, and so perform few-shot classification without unifying the support representations. None of these approaches are designed to unify partially informative support samples. The approach closest to that proposed in this paper is by \citet{edwards_towards_2017}, where the authors map the different views to a \emph{statistic} with an associated covariance through a variational approach. However there is no control of the contribution of each view to the variance, and the covariance is spherical, so the approach is also unsuitable for partial observation.

\subsection{Optimisation-Based Meta-Learning}
The few-shot classification task can also be solved without learning embeddings. One sensible baseline, fine-tuning of a previously pre-trained large model, simply treats each few-shot task as a standard classification problem \citep{nakamura_revisiting_2019}. For each task, one or more additional output layers are added on top of a pre-trained embedding network and trained to predict the classes of the support set (alongside optionally finetuning the embedding network). This can then be utilised to predict the classes of the query set.

Taking this approach a step further, Model-Agnostic Meta-Learning (MAML) \citep{finn_model-agnostic_2017} learns the initialisation of the embedding network, such that it can be rapidly fine-tuned on a new few-shot task. Given the generality of this approach, many variants of this method now exist, such as MAML++, \citet{antoniou_how_2018}, Meta-SGD \citep{li2017meta}, CAVIA \citep{zintgraf_fast_2019} and fo-Proto-MAML \citep{triantafillou_meta-dataset_2020}. One variant, LEO \citep{rusu_meta-learning_2019}, performs the meta-optimisation on a latent representation of the embedding parameters, learned using a relational network \citep{sung_learning_2018}. However, none of these variants of this fundamental optimisation based approach to few-shot learning (referred to as 'MAML' for the remainder of this work) have a mechanism for integrating partial information from the entire support set at inference time, or for comparison with a partial query observation.

\subsection{Other Meta-Learning Approaches}
Probabilisitic meta-learning methods, such as VERSA \citep{gordon_meta-learning_2019}, DKT \citep{patacchiola_bayesian_2020} and Amortised Bayesian Prototype Meta-Learning \citep{sun_amortized_2021}, often unify both embedding-based and optimisation based meta-learning by learning to output a posterior distribution that captures uncertainty in predictions, but do not use uncertainty in features to optimally combine support set information. Other recent work, such as DeepEMD \citep{zhang_deepemd_2022}, has considered the use of attention mechanisms or transformers with image patches \citep{hiller_rethinking_2022, dong_learning_2020}, or augmentations \citep{chen_few-shot_2021}. However, the purpose of these approaches is to identify informative patches or features \textit{within} each support example, to improve fine-grained few-shot learning performance or interpretability where relevant features may occupy only a small region of the samples. As far as we know, there are no existing meta-learning methods that aim to integrate partial information from \textit{across} the support set for comparison with a partially informative query.

\subsection{Partial Observability and Product of Experts}
Factor analysis is the linear counterpart to modern representation learners, but where partial observability is inherently expressed in the model. The inferential model for the latent space in factor analysis is a product of each of the conditional Gaussian factors. In general, this form of inferential model can be captured as a product of experts \citep{hinton2002training}. When those experts are Gaussian distributions~\citep{williams_prod_gaussians}, this product of experts is fully tractable. By focusing on the inferential components rather than the linear model, it is possible to generalise factor analysis inference to nonlinear mappings \citep{tang_deep_2012}. However, when only an inferential component is required (as with representation learning), the product of experts can be used more flexibly, as in our approach below.

\section{Theoretical Formalism}\label{sec:theory}
In this section, we introduce POEM, which incorporates a product of experts model for combining different views with a prior representation, and then uses that representation to classify a query view.

\subsection{Product of Expert Prototypes}
Let us consider data corresponding to partial observations, or views, of a set of items. In common with most few-shot frameworks, we arrange the data into support sets and query sets. Each support set consists of $M$ data items: $S=\{\XB^{m}|{m=1,2,\ldots,M}\}$, where the $m$th item $\XB^{m}$ collects $V^m$ views, where $V$ may vary with $m$. Let $\xB^m_v$ denote the $v$th view of the $m$th data item, such that $\XB^m=\{\xB^m_1,\xB^m_2,\ldots,\xB^m_{V^m}\}$. The items in the support set are sampled randomly from the training dataset. The query point, denoted $\xB^*$, here consists of a single \emph{different} view corresponding to one and only one of the $M$ items in the support set (although in general we may consider $N$ query points simultaneously). We cast our representation learning problem as a meta-learning task. We must learn a unified representation derived from the support set that can be compared with a representation of the query view. We want that comparison to enable us to infer which support set item $m=m^*$ the query view belongs to.

In this paper we are concerned with \emph{partial observability}; that is, not every data view will inform the whole representation. So instead of mapping each view to a deterministic point representation, we map each view to a distributional representation where each component is a normalised density that indicates the uncertainty in that component (called a \textit{factor}). We denote this conditional density $\phi$, and on implementation parameterise the parameters of the distribution $\phi$ with a neural network. We combine the corresponding factors for each view together using a product of experts, which integrates a prior distribution along with the different views such that views with low variance in a component strongly inform that component.

For a given support set, we compute a product of experts distribution for the representation $\zB^m$:
\begin{equation}
p(\zB^{m}|\XB^{m})=\frac{p(\zB^{m})\prod_{v=1}^{V^{m}}\phi(\zB^{m}|\xB_v^{m})}{\int d\zB^{\prime}\ p(\zB^{\prime})\prod_{v=1}^{V^{m}}\phi(\zB^{\prime}|\xB_v^{m})}, \label{eqn:support}
\end{equation}
where $p(\zB)$ is a prior density over the latent space. Now for a query point with a view that matches other views from e.g.\ data item $m$, we can use Bayes rule to compute the probability that the query point would be generated from the corresponding representation $\zB^m$ by
\begin{equation}
p(\xB^*|\zB^{m})=\frac{p(\xB^*)\phi(\zB^{m}|\xB^*)}{p(\zB^{m})},\label{eqn:query}
\end{equation}
where, again, $p(\zB)=\int d\xB \ p(\xB)\phi(\zB|\xB)$ is the prior.

We put Eq.\ref{eqn:query} and Eq.\ref{eqn:support} together and marginalise over $\zB^{m}$ to get the marginal predictive distribution
\begin{align}
p(\xB^*|\XB^{m})&=\int d\zB^{m}\ \left(\frac{p(\zB^{m})\prod_{v=1}^{V^{m}}\phi(\zB^{m}|\xB_v^{m})}{\int d\zB^{\prime}\ p(\zB^{\prime})\prod_{v=1}^{V^{m}}\phi(\zB^{\prime}|\xB_v^{m})}\right)\left(\frac{p(\xB^*)\phi(\zB^{m}|\xB^*)}{p(\zB^{m})}\right)\\
&=p(\xB^*) \left(\frac{\int d\zB^{m}\ \phi(\zB^m|\xB^*)\prod_{v=1}^{V^{m}}\phi(\zB^{m}|\xB_v^{m})}{\int d\zB^{\prime}\ p(\zB^{\prime})\prod_{v=1}^{V^{m}}\phi(\zB^{\prime}|\xB_v^{m})}\right)
=p(\xB^*) \frac{\lambda(\xB^*, \XB^m)}{\lambda^{\prime}(\XB^m)}\label{eqn:query|support}
\end{align}
where
\begin{align}
    \lambda(\yB, \XB)&=\int d\zB\ \phi(\zB|\yB)\prod_{v=1}^{V}\phi(\zB|\xB_v),\hspace{3mm} \mbox{ and}\label{eqn:lambda}\\
    \lambda^{\prime}(\XB)&=\int d\zB\ p(\zB)\prod_{v=1}^{V}\phi(\zB|\xB_v).\label{eqn:lambdaprime}
\end{align}

The marginal predictive $p(\xB^*|\XB^{m})$ is used to form the training objective. In our few shot task, we wish to maximize the likelihood for the correct match of query point to support set, accumulated across all support/query selections indexed with $t$ from the dataset. This provides a complete negative log marginal likelihood objective to be minimized, as derived in appendix \ref{app:objderivation}:
\begin{equation}
\mathcal{L}(\{S_t\},\{x_t^*\})=-\sum_t \left[\log \frac{\lambda(\xB^*, \XB^{m^*})}{\lambda^{\prime}(\XB^{m^*})} - \log \sum_{m}\frac{\lambda(\xB^*, \XB^m)}{\lambda^{\prime}(\XB^m)}\right]\label{eqn:objective}
\end{equation}

Full pseudocode for training POEM with this objective is provided in appendix \ref{app:pseudocode}.

\subsection{Interpretation of Objective}\label{Interpretation}
While the normalised factors $\phi$ can be chosen from any distribution class, we take $\phi$ to be Gaussian with parameterised mean and precision for the remainder of this paper, rendering the integral in Eq.~\ref{eqn:lambda} analytic. Approximating the prior $p(\zB)$ by a Gaussian also renders Eq.~\ref{eqn:lambdaprime} analytic. \footnote{In reality, $p(z)$ is typically flat over the region of non-negligible density of the product $\prod_{v=1}^{V}\phi(\zB|\xB_v)$ so does not affect the value of $\lambda^{\prime}$ in Eq.~\ref{eqn:lambdaprime} and can be neglected, as described in appendix \ref{app:GaussianProductRules}.} We note that other distributions with analytic products, such as Beta distributions, may also be of interest in certain applications, but we leave an investigation of other distributional forms for $\phi$ to further work.

If the representations from each view for a support point are aligned with each other and the query view (the means of all the Gaussians are similar), they will have a greater overlap and the integral of the resulting product of Gaussians will be larger, leading to a greater value of $\lambda(y,\XB)$. Furthermore, increasing the precisions for aligned Gaussian components leads to greater $\lambda(y,\XB)$, while, up to a limit, decreasing the precisions for non-aligned Gaussian components leads to greater $\lambda(y,\XB)$.

While the numerator in Eq.~\ref{eqn:query|support}, $\lambda(y,\XB)$, quantifies the overlap of the support set with the query, the denominator $\lambda^{\prime}(\XB)$ contrasts this with the overlap of the support set representation with the prior. Together, this factor is enhanced if it is beneficial in overlap terms to replace the prior with the query representation, and reduced if such a replacement is detrimental.
A greater consistency between query and combined support set representations intuitively leads to a greater probability that the query belongs to the class of the corresponding support set, effectively extending Prototypical Networks to a probabilistic latent representation space~\citep{snell_prototypical_2017}.

As a result, this objective is a generalisation of a Prototypical Network that allows for (a) learnable weighted averaging over support examples based on their informativeness to a given component; (b) learnable combinations of features from subsets of support examples (via differing relative precisions of components within support representations), and (c) partial comparison of the query sample with the support samples (via differing relative precisions within the query). With all precisions fixed to 1, this approach reproduces Prototypical Networks, neglecting small differences in scaling factors that arise with varying numbers of views. This relationship is derived in Appendix~\ref{app:ProtoEquivalence}.

\section{Experimental Evaluation}
There is a lack of established benchmarks specifically targeted at the evaluation of representation learning under partial observability. To design a comprehensive benchmark for few-shot representation learning under partial observability, we leverage Meta-Dataset \citep{triantafillou_meta-dataset_2020}, a recently proposed collection of few-shot learning benchmarks. We selected Meta-Dataset as the basis for our adapted benchmark as it consists of diverse datasets involving natural, human-made and text-based visual concepts, with a variety of fine-grained classification tasks that require learning from varying and unbalanced numbers of samples and classes. As a result, our derived benchmark inherits these properties to provide a robust measure of the ability of a learning approach to learn representations from partial observations.

\begin{figure}[h!]
  \centering
  \includegraphics[width=0.9\textwidth]{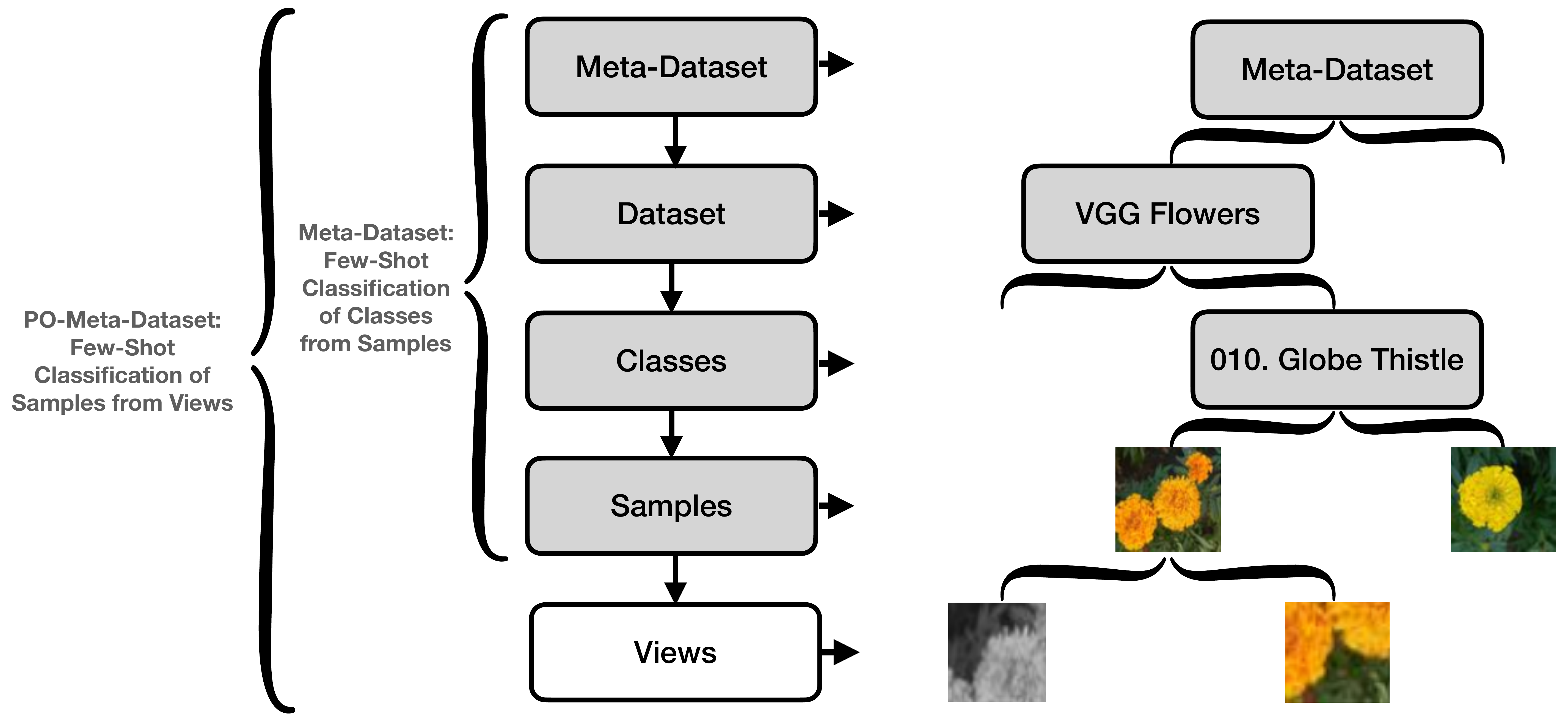}
  \caption[MMD caption]%
  {Standard few-shot learning requires the prediction of an image class from a sample. Our adapted task evaluates representation learning under partial observability by instead requiring prediction of the underlying image from partial views. Views are generated with the standard contrastive augmentations, with stronger cropping. We call the resulting benchmark Partially Observable Meta-Dataset (PO-Meta-Dataset).}
\label{fig:MMD}
\end{figure}

To extend Meta-Dataset to incorporate partial observability, we take multiple views of \textit{each} sample and divide these views into support and query sets. Our adapted few-shot classification task is to predict which sample a query view comes from, given a selection of support views of that sample, as demonstrated in Figure~\ref{fig:MMD}. 

In keeping with the spirit of Meta-Dataset, we vary the number of ways in the task (now the number of images) from 5 to 25, taken from between 1 to 5 classes. Views are generated by applying the standard augmentation operations used in SimCLR \citep{chen_simple_2020} and most other self-supervised learning methods. However, to emphasise the focus on partial observability, the size of the random crops and the number of views was fixed, such that the entire support set for a sample contains a maximum of 50\% of the image. We also maintain a constant number of query views per sample. Viewpoint information consisting of the coordinates of the view is provided to make it possible for learners to understand where a view fits into a representation even in the absence of overlapping views. Full details of the definition of the task are provided in appendix \ref{app:MDDetails}.

We apply our proposed evaluation procedure to all datasets included in Meta-Dataset with a few exceptions. ILSVRC (ImageNet, \citet{russakovsky_imagenet_2015}) was not included since our network backbones were pre-trained on this dataset, including the standard few-shot test classes (which is also why this dataset was subsequently removed from the updated benchmark, MetaDataset-v2 \citep{dumoulin_comparing_2021}). Traffic Signs \citep{stallkamp2011german} and MSCOCO \citep{lin_microsoft_2015} were not included since these datasets are fully reserved for evaluation by Meta-Dataset and so do not have a training set specified. Quick Draw \citep{fernandez-fernandez_quick_2019} was also not included since this dataset was found to be too large to use within the memory constraints of standard RTX2080 GPUs. This leaves six diverse datasets: Aircraft \citep{maji_fine-grained_2013}, Birds \citep{wah2011caltech}, Flowers \citep{Nilsback08}, Fungi \citep{schroeder_brigit_fgvc5_2018}, Omniglot \citep{lake2015human} and Textures \citep{cimpoi2014describing}, on all of which our models were trained, validated and tested on according to the data partitions specified by the Meta-Dataset benchmark.

The resulting benchmark, Partially Observed Meta-Dataset (PO-Meta-Dataset), therefore requires that the learner coherently combine the information from the support views into a consistent representation of the sample, such that the query view can be matched to the sample it originated from. Since a maximum of 50\% of each sample is seen in the support set, the task also requires generalisation to correctly match and classify query views.

\subsection{Implementation Details}
We utilise a re-implementation of Meta-Dataset benchmarking in PyTorch \citep{paszke_pytorch_2019} which closely replicates the Meta-Dataset sampling procedure of uniformly sampling classes, followed by a balanced query set (since all classes are considered equally important) and unbalanced support sets (to mirror realistic variations in the appearances of classes). The experimental implementation, including full open-source code and data will be available on publication.

Following the \textit{MD-Transfer} procedure used in Meta-Dataset, we leverage a single ResNet-18 \citep{he_deep_2015} classifier pre-trained on ImageNet \citep{russakovsky_imagenet_2015} at $126 \times 126$ resolution. Since both a mean and precision must be learned to fully specify the model $\phi_v(\zB|\xB_v^n)$, we add two simple 3-layer MLP heads onto this backbone for POEM, each maintaining an embedding size of $512$. For fair comparison, we also add the same 3-layer MLP head onto the backbone for the baselines. Using a larger embedding for the baselines was not found to be beneficial. During training, gradients are backpropagated through the entire network such that both the randomly initialised heads and pre-trained backbones are learned/fine-tuned.

We use a representative selection of meta-learning baselines utilised by Meta-Dataset for our re-implementation. This includes a strong naive baseline (Finetuning, \citet{nakamura_revisiting_2019}), an embedding-based approach (Prototypical Network, \citet{snell_prototypical_2017}) and an optimisation-based approach (MAML, \citet{finn_model-agnostic_2017}), all modernised to use the ResNet-18 backbone as described above. Recent competitions, such as the NeurIPS 2021 MetaDL Challenge \citep{baz_lessons_2022, baz_advances_2021}, have demonstrated that these fundamental approaches, updated to use modern pre-trained backbones that are finetuned on the meta-task (exactly as in our experiments below) are still generally state-of-the-art for novel datasets \citep{chen_metadelta_2021}, and so form strong baselines. In addition, our re-implementation enables us to ensure that all learners are optimised for Meta-Dataset and that comparisons between learners are fair, utilising the same benchmark parameters, model architectures and where applicable, hyperparameters. Crucially, given the close connection between POEM and Prototypical Networks, we ensure that all hyperparameters, including learning rates, scheduling and architectures are identical for both methods.

\subsection{Results}
Our results on this novel representation learning benchmark, PO-Meta-Dataset, are given in table \ref{tab:results1}.

\begin{table}[h!]
\centering
\begin{tabular}{| c | c | c | c | c |} 
 \hline
 Test Source & Finetune & ProtoNet & MAML & POEM \\ 
 \hline\hline
 Aircraft & $46.5\pm0.6$ & $48.5\pm1.0$ & $37.5\pm0.3$ & $\mathbf{55.3\pm0.7}$ \\ 
 \hline
 Birds & $62.6\pm0.7$ & $67.4\pm1.2$ & $52.5\pm0.6$ & $\mathbf{71.1\pm0.1}$ \\ 
 \hline
 Flowers & $48.5\pm0.4$ & $46.4\pm0.7$ & $33.5\pm0.3$ & $\mathbf{49.2\pm1.5}$ \\ 
 \hline
 Fungi & $61.0\pm0.2$ & $61.4\pm0.4$ & $46.1\pm0.4$ & $\mathbf{64.8\pm0.3}$ \\ 
 \hline
 Omniglot & $71.3\pm0.1$ & $87.8\pm0.1$ & $47.4\pm1.0$ & $\mathbf{89.2\pm0.7}$ \\ 
 \hline
 Textures & $\mathbf{83.2\pm0.4}$ & $76.7\pm1.6$ & $73.1\pm0.4$ & $81.4\pm0.6$ \\ 
 \hline
\end{tabular}
\caption{Few-shot classification accuracies on our adapted Meta-Dataset benchmark, PO-Meta-Dataset. All learners use a ResNet-18 model pre-trained on ImageNet, with MLP heads to incorporate view information. POEM outperforms the baselines across the range of datasets, demonstrating the benefits of the approach to learn and match representations from partial observations.}
\label{tab:results1}
\end{table}

The results show that POEM outperforms the baselines at identifying views of images across a diverse range of datasets, demonstrating the benefits of the approach to learn and match representations from partial observations. The only exception is the Textures dataset, for which the finetuning baseline performs particularly strongly. We hypothesise that this is because the images in the Textures dataset are relatively uniform compared to the other datasets, so capturing the relative location of views is less important than identifying very fine grained features that distinguish the samples, which optimisation-based approaches are particularly effective at.

\subsection{Ablation: Meta-Dataset}\label{Ablation}
To demonstrate that the observed benefit of POEM over the baselines is due to the requirement of the task to learn coherent representations from partial observations, we also evaluate our approach against the baselines on the established Meta-Dataset benchmark. We now follow the standard few-shot learning procedure as described in the paper \citep{triantafillou_meta-dataset_2020}, but keep all learners identical to those used in the evaluation above.

\begin{table}[h!]
\centering
\vspace*{-5mm}
\begin{tabular}{| c | c | c | c | c |} 
 \hline
 Test Source & Finetune & ProtoNet & MAML & POEM \\ 
 \hline\hline
 Aircraft & $56.2\pm1.1$ & $47.2\pm1.2$ & $35.9\pm1.8$ & $46.5\pm1.5$ \\ 
 \hline
 Birds & $52.6\pm1.8$ & $78.3\pm0.5$  & $65.2\pm0.3$ & $79.4\pm0.3$ \\ 
 \hline
 Flowers & $80.1\pm2.0$ & $84.2\pm0.7$ & $70.4\pm0.4$ & $83.6\pm1.3$ \\ 
 \hline
 Fungi & $33.6\pm1.7$ & $84.7\pm0.2$ & $18.9\pm0.2$ & $81.0\pm0.1$  \\ 
 \hline
 Omniglot & $89.6\pm3.3$ & $98.7\pm0.1$ & $94.7\pm0.1$ & $98.6\pm0.1$ \\ 
 \hline
 Textures & $60.4\pm1.0$ & $65.3\pm1.2$ & $56.1\pm0.3$ & $65.7\pm0.8$ \\ 
 \hline
\end{tabular}
\caption{Few-shot classification accuracies on Meta-Dataset, all using a ResNet-18 backbone pre-trained on ImageNet, with a 3 layer MLP head. POEM is comparable with the baselines.}
\label{tab:results2}
\end{table}

Our results on the standard Meta-Dataset benchmark are provided in table \ref{tab:results2}. As expected, we find that POEM performs comparably with the baselines. Although Meta-Dataset provides realistic few-shot learning tasks in terms of diversity of visual concepts, fine-grained classes and variable shots and ways, each sample generally contains complete information including all relevant features for the visual concept in question. Correctly classifying query samples does not generally require any unification of views from support examples, but simply the identification of common features. As a result, we see that the additional capacity of POEM to learn to weight support examples and combine partial features does not provide a significant performance improvement over the baselines at few-shot classification in this fully observable benchmark. 

In support of our hypothesis that feature uncertainty is not useful on this benchmark, we find that the variance in the precisions relative to the means output by the POEM model generally decreases during training and becomes negligible for all datasets, indicating that the precisions are not being utilised to improve performance and that the POEM objective is reducing to the Prototypical Network objective, as discussed in section \ref{Interpretation}. This is further evidenced by the very similar performances of POEM and the Prototypical Network across the entire set of datasets. However, on PO-Meta-Dataset, we find that the relative variance in the precisions to the means is much larger on convergence, which leads to the improved performance of POEM over the Prototypical Network observed in Table \ref{tab:results1}. This is shown in appendix \ref{app:Precisions}.

\section{Demonstration of Learning Representations of Environments}
We now apply POEM to the equivalent task of learning a representation of an environment from the partial observations collected by an agent exploring that environment. 

To do so, we utilise the 2D gridworld environment, MiniGrid \citep{gym_minigrid}. We consider the $11\times11$ Simple Crossing environment, which consists of a procedurally generated maze where the agent is required to traverse from the top left corner to the goal in the bottom right corner. The MiniGrid environment provides an agent-centric viewpoint at each step in the trajectory, consisting of a $7\times7$ window of the environment in front of the agent, taking into account the agent's current direction and where the line of sight is blocked by walls. 

\subsection{Meta-Learning Environment Representations via Few-Shot Classification}

To generate few-shot episodes, we utilise two agents: an optimal agent that takes the optimal trajectory from the start to the goal, and an exploratory agent that is incentivised to explore all possible views in the environment. The support set for each environment is generated by running the optimal agent in the environment and collecting the partial observations of this agent at each step during its trajectory. The query set is similarly generated by running the exploratory agent in the environment, filtering out any observations that are contained within the support set, and then randomly sampling the desired number of queries from the remaining observations. 

\begin{figure}[h!]
  \vspace*{-1mm}
  \centering
  \begin{minipage}[t]{0.39\textwidth}
  \vspace{0pt}
    \includegraphics[width=\textwidth, align=t]{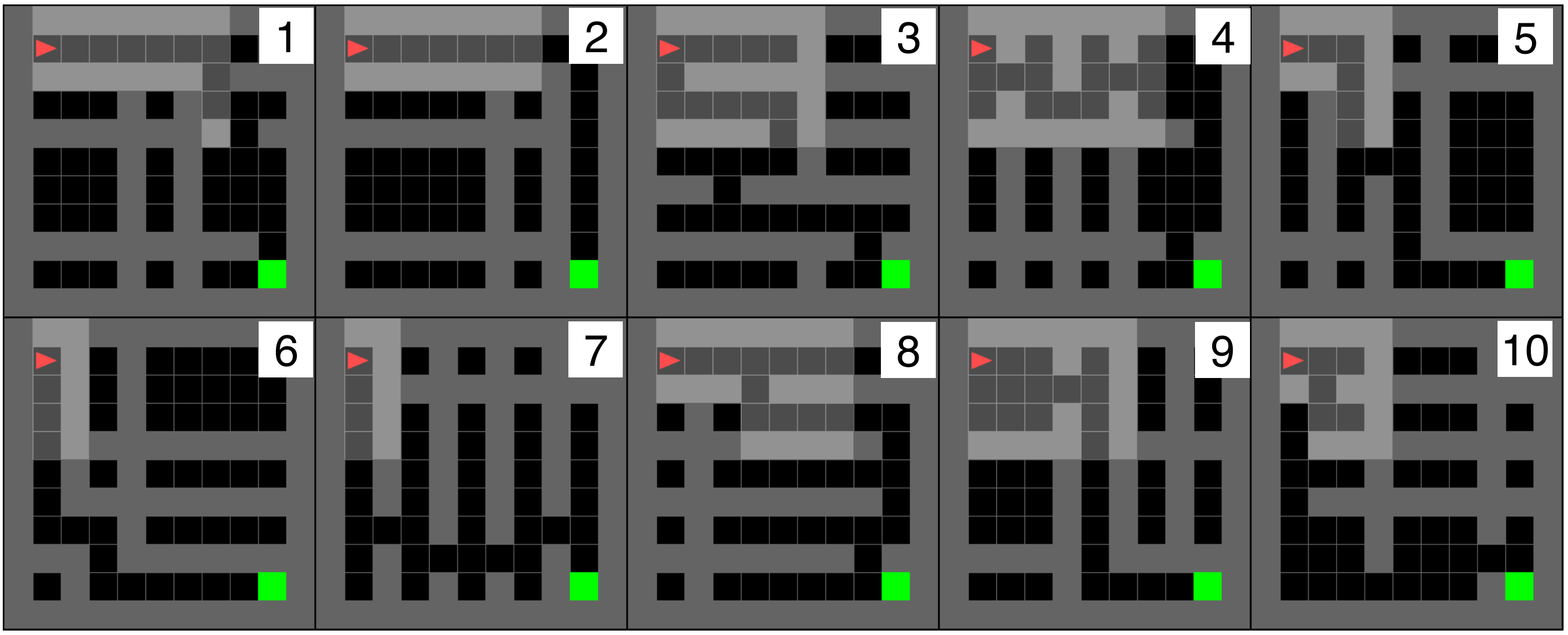}
    \caption{Sample environments.}\label{fig:SampleEnvs}
  \end{minipage}
  \hfill
  \begin{minipage}[t]{0.59\textwidth}
    \includegraphics[width=\textwidth, align=t]{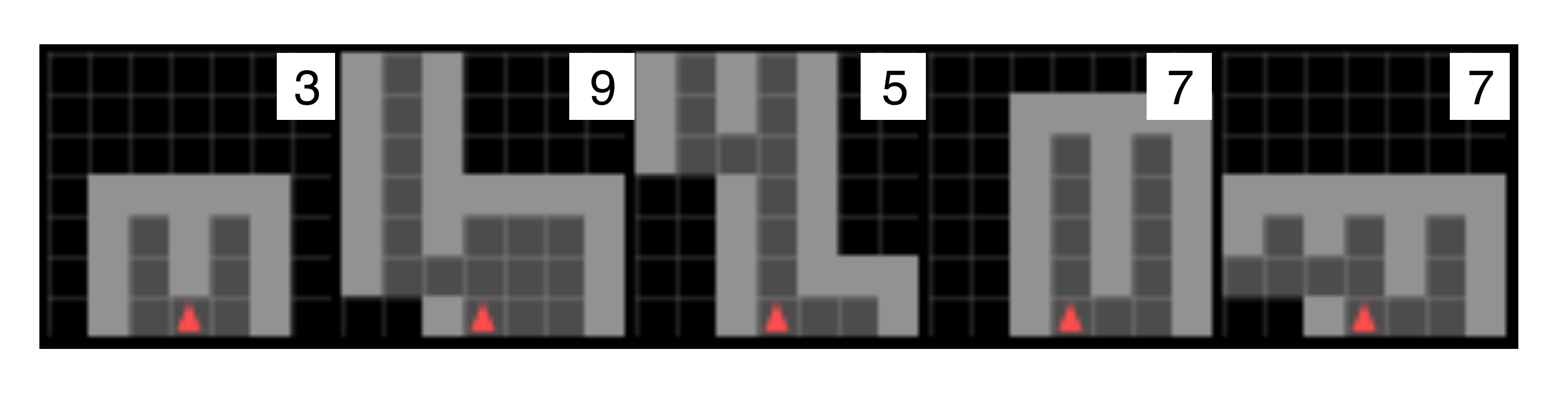}
    \caption{Sample queries labelled with targets corresponding to the environment which they were observed in.}\label{fig:SampleQueries}
  \end{minipage}
\end{figure}

We generate these few-shot episodes dynamically, and train POEM to combine the support samples (partial observations from the optimal trajectory) into a representation of the environment, such that it can classify which environment a novel query observation has been collected from. A set of sample environments and observations from those environments are shown in figures \ref{fig:SampleEnvs} and \ref{fig:SampleQueries}.

All observations are provided as pixels to a standard convolutional backbone, with the corresponding agent location and direction appended to this representation and passed through an MLP head, equivalent to the procedure utilised for the adapted Meta-Dataset experiments. As a baseline comparison, we also train a Prototypical Network with an identical architecture on this task. Additionally, we train an equivalent recurrent network architecture typically applied to POMDP tasks such as this \citep{hausknecht_deep_2017}, by adding a GRU layer \citep{cho_properties_2014, chung_empirical_2014} where the hidden state of the GRU is updated at each timestep and then extracted as the unified representation of the agent. We find that POEM trains more quickly and reaches almost 10\% higher final environment recognition performance than both the Prototypical Network and GRU-based approach over 100 test episodes (81.1\% vs 72.4\% and 72.1\%), as shown in appendix \ref{app:EnvAccuracy}. This is a result of POEM's capacity to associate each observation with only part of the representation.

\subsection{Reconstructing Environments from Partial Observation Trajectories}
\begin{figure}[h!]
  \centering
  \begin{minipage}[t]{0.49\textwidth}
    \includegraphics[width=\textwidth]{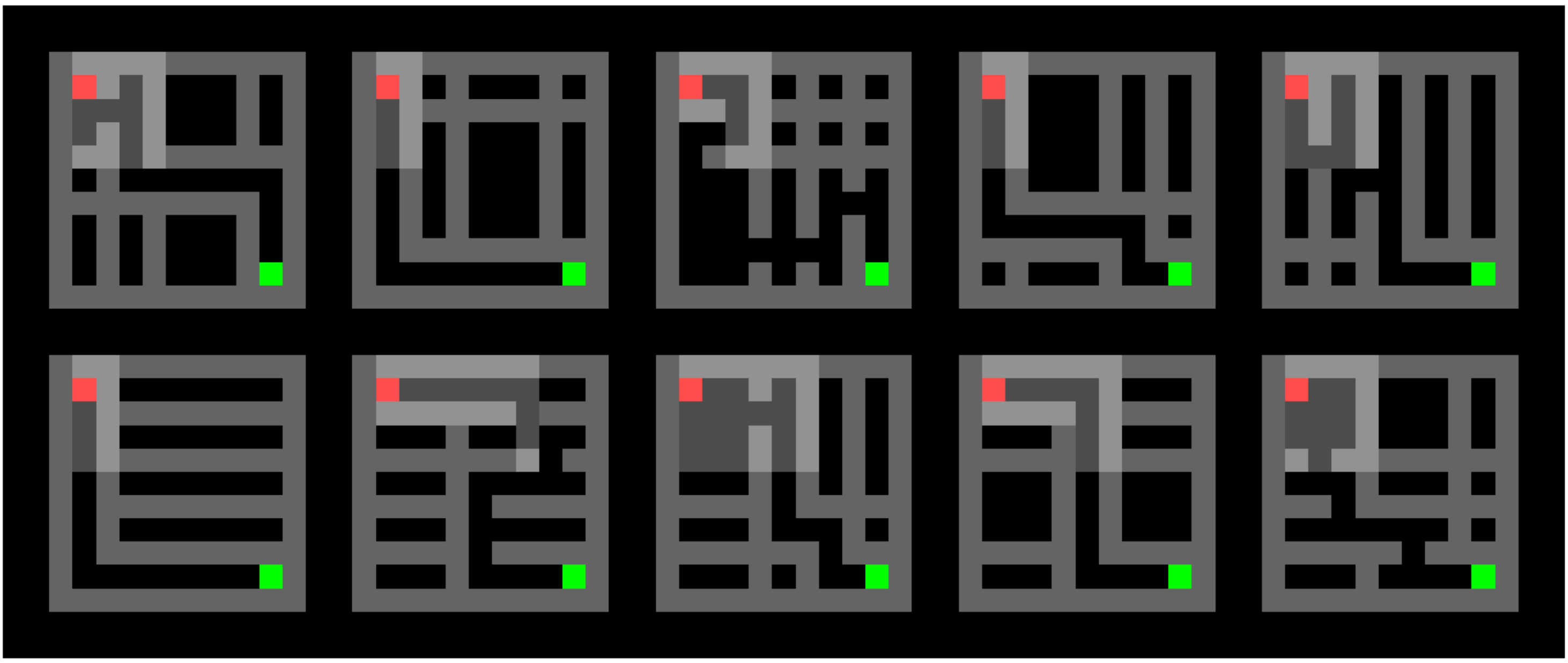}
  \end{minipage}
  \hfill
  \begin{minipage}[t]{0.49\textwidth}
    \includegraphics[width=\textwidth]{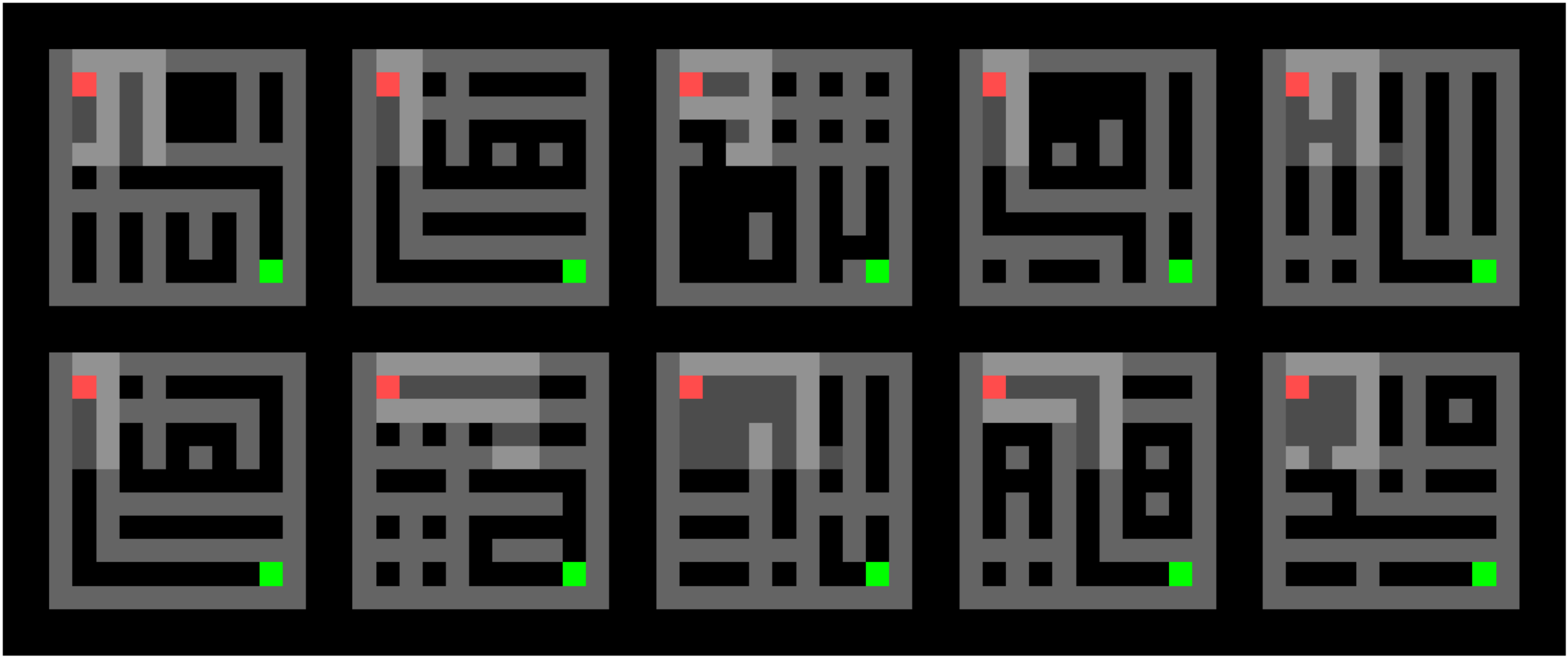}
  \end{minipage}
  \caption{\textbf{Left}: Ground truth environments explored by the agent. \textbf{Right}: Reconstructions of the corresponding environments from POEM's unified representation, encoded from the partial observations of the agent.}\label{fig:Reconstructions}
\end{figure}

Having learned an environment encoder using the few-shot learning procedure above, we now investigate the extent to which our representations can be used to reconstruct the environment. As above, we generate trajectories with the optimal agent and feed these through the encoder to generate a representation of the environment. An MLP decoder is then trained to reconstruct the original environment layout from the learned environment representation. The decoder attempts to predict a one-hot representation of each possible grid cell, with a mean squared error loss. Given the trained encoder and decoder, we are now able to generate a map of the environment the optimal agent has traversed, solely from the agent's partial observations, and without ever having seen the environment as a whole. A sample of environments alongside their reconstructions are shown in figure \ref{fig:Reconstructions}.

We see that the reconstructions clearly capture the approximate structure of each environment, demonstrating that the agent has been able to integrate its observations from along its trajectory into a single consistent representation. Since POEM enables the representation to be updated incrementally with each partial observation of the environment at inference time, it would be possible for an agent to update an internal environment representation at each step in its trajectory. There is potential for utilising this approach for learning environment representations to be beneficial in the context of exploration for reinforcement learning, but we leave such an investigation to future work.

\section{Conclusion}
In this work, we have introduced Partial Observation Experts Modelling (POEM), a contrastive meta-learning approach for few-shot learning in partially-observable settings. Unlike other standard contrastive and embedding-based meta-learning approaches, POEM utilises representational uncertainty to enable observations to inform only part of a representation vector. This probabilistic formalism enables consistent representation learning from multiple observations with a few-shot learning objective. We have demonstrated that POEM is comparable to the state-of-the-art baselines on a comprehensive few-shot learning benchmark, and outperforms these baselines when this benchmark is adapted to evaluate representation learning from partial observations. We have also demonstrated a promising potential application for POEM to learn representations of an environment from an agent's partial observations. We hope that this research inspires further work into the challenging task of learning representations under partial observability and the creation of more realistic partial observability benchmarks.
\newpage

 \subsubsection*{Acknowledgments}
Adam Jelley was kindly supported by Microsoft Research and EPSRC through Microsoft's PhD Scholarship Programme. Antreas Antoniou was supported by a Huawei DDMPLab Innovation Research Grant. The Meta-Dataset experiments in this work were partly funded by Google Research Compute Credits, and we thank Hugo Larochelle for his support in acquiring these compute credits.


\bibliography{references,amosreferences}
\bibliographystyle{iclr2023_conference}

\appendix
\section{Appendix}
\subsection{Gaussian Product Rules}\label{app:GaussianProductRules}
Assuming the latent variable model $\phi(\zB|\xB)$ to be a diagonal covariance multivariate Gaussian, the resulting integrals over latent variables become integrals over Gaussian products. This allows both $\lambda(\yB,\XB)$ (Equation \ref{eqn:lambda}) and $\lambda^\prime(\XB)$ (Equation \ref{eqn:lambdaprime}) in the marginal predictive distribution (Equation \ref{eqn:query|support}) to be evaluated analytically using the following univariate Gaussian product rules on each independent dimension \citep{roweis_sam_gaussian_1999}.

Since a product of Gaussians $\prod_i N(\mu_i,\tau_i^{-1})$ is itself a Gaussian, we have $\prod_i N(\mu_i,\tau_i^{-1}) = S N(\mu,\tau^{-1})$, where
\begin{align}
    \tau &= \sum_i \tau_i \\
    \mu&=\frac{1}{\tau}\sum_i \tau_i \mu_i \\
    S&= (2\pi)^{\frac{(1-n)}{2}} \frac{\prod_i \tau_i^{1/2}}{\tau^{1/2}}\exp\left(\frac{1}{2}\tau \mu^2 - \frac{1}{2} \sum_i \tau_i \mu_i^2\right).
\end{align}
Therefore the integral of a Gaussian product is given by the resulting normalisation $S$.

In the case of evaluating the marginal predictive distribution $p(\xB^*|\XB^{m})$ (equation \ref{eqn:query|support}), this gives $\frac{S^*S}{S'S}=\frac{S^*}{S'}$ where $S$ is the normalisation constant of the support product, $S^*$ is the normalisation constant of the product of the query and normalised support product, and $S'$ is the normalisation constant of the product of the prior $p(z)$ and normalised support product. In reality, $p(z)$ generally has little impact as it is typically flat ($\tau \rightarrow 0$) over the region of non-negligible density of the product $\prod_{v=1}^{V}\phi(\zB|\xB_v)$ and so $S'\approx 1$ and we find $\frac{S^*}{S'} \approx S^*$ so the ratios $\frac{\lambda}{\lambda'}$ in the objective can be approximated by $S^*$, as in the simplified pseduocode in appendix~\ref{app:pseudocode}.

\subsection{Derivation of Objective from Marginal Predictive Distribution}\label{app:objderivation}
In section \ref{sec:theory}, we derived the marginal predictive distribution:
\begin{align}
p(\xB^*|\XB^m)=p(\xB^*) \frac{\lambda(\xB^*, \XB^m)}{\lambda^{\prime}(\XB^m)}
\end{align}
where
\begin{align}
    \lambda(\yB, \XB)&=\int d\zB\ \phi(\zB|\yB)\prod_{v=1}^{\mathrm{dim}(\XB)}\phi(\zB|\xB_v),\hspace{3mm} \mbox{ and}\\
    \lambda^{\prime}(\XB)&=\int d\zB\ p(\zB)\prod_{v=1}^{\mathrm{dim}(\XB)}\phi(\zB|\xB_v).
\end{align}
In our few shot task, the support data is chosen and then the query view is chosen uniformly at random to match the views of one of the support data items. Let the hypothesis $H_{m}$ indicate the event that the query view $\xB^*$ comes from support point $m$. Then 
\begin{align}
P(H_{m}|S, \xB^*)&=\frac{P(H_{m})P(S, \xB^*|H_{m})}{\sum_{m^\prime} P(H_{m^\prime})P(S, \xB^*|H_{m^\prime})}=\frac{(1/M)p(\xB^*|S,H_m)}{\sum_{m^\prime}(1/M)p(\xB^*|S,H_{m^\prime})}\\&=\frac{(1/M)p(\xB^*|\XB^{m})}{\sum_{m^\prime}(1/M)p(\xB^*|\XB^{m^\prime})}=\frac{p(\xB^*|\XB^{m})}{\sum_{m^\prime} p(\xB^*|\XB^{m^\prime})}.\label{eqn:pH}
\end{align}
From this we can formulate the training task: we wish to maximize the likelihood for the correct match of query point to support set, accumulated across all support/query selections from the dataset. Denote the $t$th support set by $S_t$, the $t$th query point by $x_t^*$, and let $m_t$ denote the support point with views that match the view of the query point. Then the complete negative log marginal likelihood objective to be minimized is:
\begin{align}
\mathcal{L}(\{S_t\},\{x_t^*\})&=-\sum_t \log P(H_{m_t}|S_t, \xB_t^*)\\
&=-\sum_t \log \frac{p(\xB^*|\XB^{m})}{\sum_{m^\prime} p(\xB^*|\XB^{m^\prime})}\\
&= -\sum_t \left[\log \frac{\lambda(\xB^*, \XB^{m^*})}{\lambda^{\prime}(\XB^{m^*})} - \log \sum_{m}\frac{\lambda(\xB^*, \XB^m)}{\lambda^{\prime}(\XB^m)}\right]
\end{align}

\newpage
\subsection{Pseudocode}\label{app:pseudocode}
\begin{algorithm}[th!]
\caption{Pytorch-Style Pseudocode: Gaussian Partial Observation Experts Modelling}
\label{alg:code}
\definecolor{codeblue}{rgb}{0.25,0.5,0.5}
\definecolor{codekw}{rgb}{0.85, 0.18, 0.50}
\lstset{
  backgroundcolor=\color{white},
  basicstyle=\fontsize{7.5pt}{7.5pt}\ttfamily\selectfont,
  columns=fullflexible,
  breaklines=true,
  captionpos=b,
  commentstyle=\fontsize{7.5pt}{7.5pt}\color{codeblue},
  keywordstyle=\fontsize{7.5pt}{7.5pt}\color{codekw},
  escapechar=ß
}
\begin{lstlisting}[language=python]
# phi: dual-headed encoder network with shared backbone and output heads for mean and precision of Gaussian embedding
# M: Number of items/classes in task
# V: Number of views of each item/class (in general can vary with m in range(M))
# N: Number of query views
# D: Embedding dimension

# Load augmented partial views with view information
for (support_views, query_views, query_targets) in loader: 
    # support_views.shape = (M, V, ...)
    # query_views.shape=(N, ...)
    # query_targets.shape = (N,) 

    # Encode each support and query views
    support_means, support_precisions = phi(support_views) # (M, V, D)
    query_means, query_precisions = phi(query_views) # (N, D)
    
    # Combine support views into unified representation of each item
    # Gaussian products computed using equations in appendix A.1
    # Optionally include prior Gaussian here (neglected for simplified implementation)
    environment_means, environment_precisions, log_environment_normalisation = inner_gaussian_product(support_means, support_precisions) # Outputs: (M, D)
    
    # Combine each query view with each unified support representation
    env_query_mean, env_query_precisions, log_env_query_normalisation = outer_gaussian_product(support_means, support_precisions, query_means, query_precisions) # Outputs: (N, M, D)
    
    # Predictions correspond to unified support with maximum overlap with query
    _, predictions = log_env_query_normalisation.sum(2).max(1) # (N,)
    
    # Cross entropy loss normalises with softmax and computes negative log-likelihood
    loss = F.cross_entropy(log_env_query_normalisation, query_targets, reduction='mean')
    
    # Optimization step
    loss.backwards()
    optimizer.step()
    
\end{lstlisting}
\end{algorithm}

\begin{algorithm}
\caption{Language Agnostic Pseudocode: Gaussian Partial Observation Experts Modelling}\label{alg:cap}
\begin{algorithmic}[1]
\Require Training meta-set $D^{train}\in\mathcal{T}$
\Require Learning rate $\alpha$
\State Initialise dual-headed network $\phi_\theta(\zB|\xB)$\\
\Comment{Heads correspond to mean $\muB$ and precision $\tauB$ of Gaussian embedding $\zB$}
\While {not converged}
\State Sample task instance  $\mathcal{T}_i = (\XB, \xB^*) \sim D^{train}$\\
\Comment{Support set $\XB$ consists of $V^m$ views of item $m \in \{1,...,M\}$.}\\
\Comment{Query set $\xB^*$ consists of $N$ queries, each one view from any one item.}
\State Encode each view in support set $\XB$ into Gaussian $\zB$ using $\phi(\zB|\XB)$
\State Encode each query view in $\xB^*$ into Gaussian $\zB^*$ using $\phi(\zB^*|\xB^*)$
\For {$m \in \{1,...,M\}$}
\State Compute Gaussian product over views $\prod_{v=1}^{V^m}\phi(\zB|\xB_v^m)$ (using results in \ref{app:GaussianProductRules})\\
\Comment{This gives unified support representation (global environment representation)}
\For{$n \in \{1,...,N\}$}
\State Compute Gaussian product of query with support product $\phi(\zB_n^*|\xB_n^*)\prod_{v=1}^{V^m}\phi(\zB|\xB_v^m)$
\EndFor
\EndFor
\State Normalise resulting query-support normalisation constants $\overline {S^{m}_n}=\frac{S^{m}_n}{\sum_m S^m_n}$ across items
\State Compute negative log of $\overline {S^{m^*}_n}$ for correct support as loss $\mathcal{L}(\{D_t\},\{x_t^*\})$ (eq. ~\ref{eqn:objective})\\
\Comment{Negative log likelihood for correct support}
\State Perform gradient step w.r.t. $\theta$: $\theta \leftarrow \phi-\alpha \nabla_\theta \mathcal{L}(\{D_t\},\{x_t^*\})$
\EndWhile
\end{algorithmic}
\end{algorithm}

\subsection{Equivalence of Prototypical Network Objective to POEM Objective with Fixed Precisions}\label{app:ProtoEquivalence}
The probability of a query $\xB^*$ belonging to class $n$ using the POEM objective is given by:
\begin{equation}
   P(H_{m}|S, \xB^*)
   = \frac{\lambda(\xB^*;\XB^{n})}{\lambda^{\prime}(\XB^{n})} \bigg/ \sum_m \frac{\lambda(\xB^*;\XB^m)}{ \lambda^{\prime}(\XB^m)}
\end{equation}
as defined in equation \ref{eqn:pH}, where
\begin{align}
    \lambda(\yB, \XB)&=\int d\zB\ \phi(\zB|\yB)\prod_{v=1}^{V}\phi(\zB|\xB_v),\hspace{3mm} \mbox{ and}\\
    \lambda^{\prime}(\XB)&=\int d\zB\ p(\zB)\prod_{v=1}^{V}\phi(\zB|\xB_v).
\end{align}.

Taking the precisions of the all Gaussian factors $\phi$ in $\lambda$ and $\lambda^{\prime}$ to be 1, we can apply the Gaussian product rules given in appendix \ref{app:GaussianProductRules} to calculate $\lambda$ and $\lambda^{\prime}$ analytically. We find that this gives:
\begin{equation}
    p_{n} = \frac{\frac{V_n}{V_n+1}^\frac{1}{2} \exp \left(-\frac{V_n}{2(V_n+1)}\left(\mu-\frac{\sum_i \mu_{ni}}{V_n}\right)^2\right)}{\sum_m \frac{V_m}{V_m+1}^\frac{1}{2}\exp \left(-\frac{V_m}{2(V_m+1)}\left(\mu-\frac{\sum_i \mu_{mi}}{V_m}\right)^2\right)}
\end{equation}
where $\mu$ is the representation mean of the query, and $\mu_{ni}$ is the representation mean of support sample $i$ for class $n$, and $V_n$ is the number of support samples for class $n$.

Equivalently, the probability of a query with representation vector $\mu$ belonging to a class $n$ using a Prototypical Network objective is given by:
\begin{equation}
    p_{n} = \frac{\exp \left(-\left(\mu-\frac{\sum_i \mu_{ni}}{V_n}\right)^2\right)}{\sum_m \exp \left(-\left(\mu-\frac{\sum_i \mu_{mi}}{V_m}\right)^2\right)}
\end{equation}
We find that these are equivalent aside from the scaling factors $\frac{V_m}{(2)(V_m+1)}$ which only have a (significant) effect when there are varying numbers of samples by class, and a greater effect when the number of samples is smaller. Experimentally, we find that these scaling factors make little difference, as demonstrated in table \ref{tab:results2} of section \ref{Ablation}.

\newpage
\subsection{PO-Meta-Dataset Benchmark Additional Details}\label{app:MDDetails}
Parameters used for adapted PO-Meta-Dataset are provided in Table \ref{tab:POMDParams}. All parameters not listed chosen to match Meta-Dataset defaults. All augmentations are applied using Torchvision, with parameters specified.

\begin{table}[h!]\label{tab:POMDParams}
\caption{PO-Meta-Dataset Parameters}
\label{sample-table}
\begin{center}
\begin{tabular}{ll}
\multicolumn{1}{c}{\bf PARAMETER}  &\multicolumn{1}{c}{\bf VALUE}
\\ \hline \\

Classes per Task         & $[1,5]$ \\
Samples per Task             & $[5,25]$ \\
Support Views per Sample    & $18$ \\
Query Views per Sample  & $2$ \\
Image Size & $(84,84)$ (except Omniglot, $(28,28)$)\\
Crop Size & $(14, 14)$ ($1/6$ in each dim, except Omniglot, random placement)\\
Color Jitter & $(0.8,0.8,0.8,0.2)$, $p(apply)=0.3$\\
Random Greyscale & $0.2$\\
Random Horizontal Flip & $0.5$\\
Gaussian Blur & $((3,3),(1,0,2.0))$, $p(apply)=0.2$
\end{tabular}
\end{center}
\end{table}

All results computed over three runs. The Finetuning, Prototypical Network and POEM baselines were run on on-premise RTX2080 GPUs. MAML required more memory and compute than available, so was run on cloud A100s.

\subsection{Relative Variance of Precisions During Training on Meta-Dataset and Meta-Meta-Dataset}\label{app:Precisions}
The plot below shows the evolution of the variance in the representation precisions relative to the variance in the representation means learned by POEM on two distinct datasets, Aircraft and VGG Flowers. We see that for standard few-shot learning on Meta-Dataset, the variance in precisions is negligible relative to the variance in the means, demonstrating that the representational uncertainty is not useful in this task. Meanwhile, we see the variance in the precisions relative to the variance in the means becoming large before converging to a value of $\mathcal{O}(100)$ on the Meta-Meta-Dataset task, demonstrating that learning relative precisions is useful in this setting since each support sample only informs part of the representation.
\begin{figure}[H]
  \centering
  \includegraphics[scale=0.5]{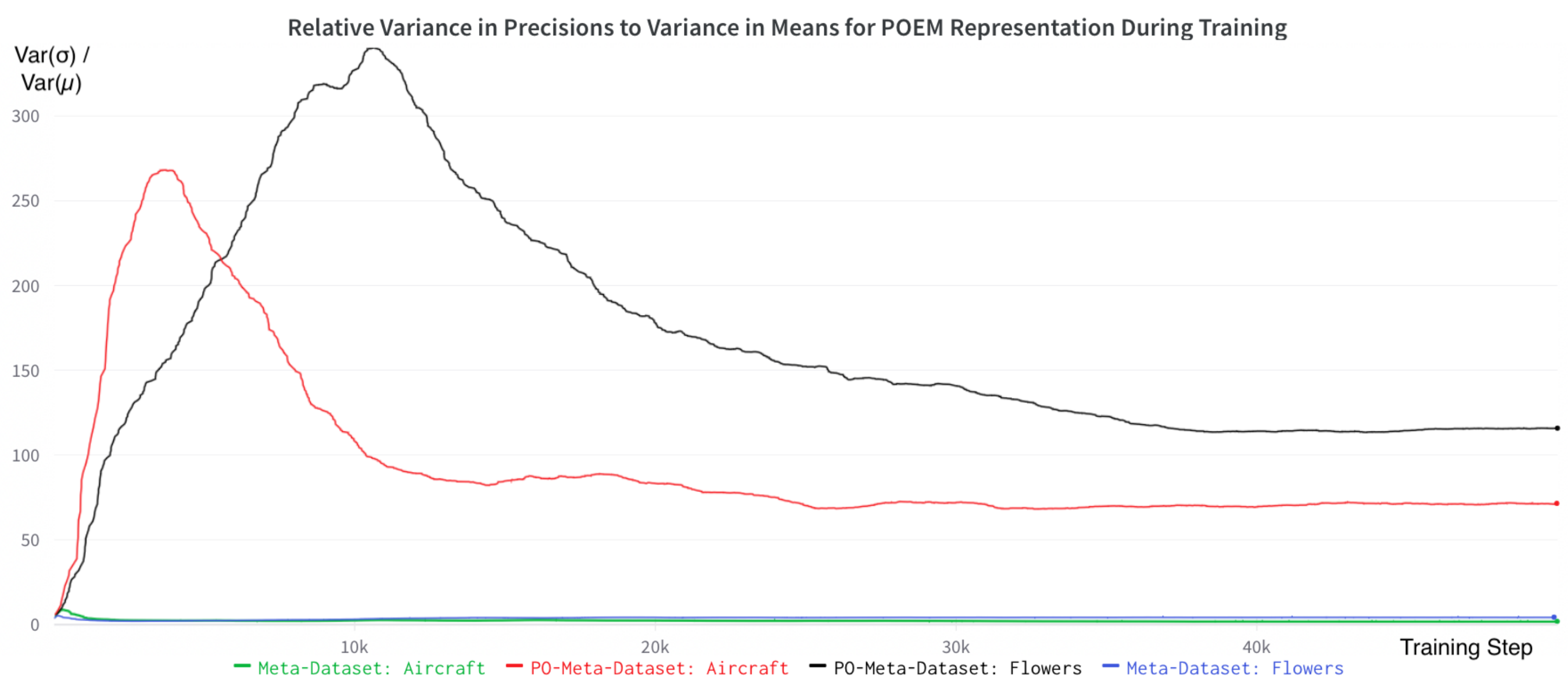}
 
\label{fig:POEM_precisions}
\end{figure}

\subsection{Learning Representations of Environments Additional Details}\label{app:EnvDetails}

Additional details about the parameters used for learning environment representations from agent observations are provided in Table \ref{tab:EnvDetails}

\begin{table}[h!]
\caption{Environment Representation Learning Parameters}
\label{tab:EnvDetails}
\begin{center}
\begin{tabular}{ll}
\multicolumn{1}{c}{\bf PARAMETER}  &\multicolumn{1}{c}{\bf VALUE}
\\ \hline \\
Agent Training Algorithm & PPO \citep{schulman_proximal_2017} (default hyperparameters)\\
Optimal Agent Reward & 1 for reaching goal, -0.01 per timestep \\
Exploratory Agent Reward & 1/N count exploration bonus (state defined by agent location and direction)\\
Encoder Conv Backbone Layers & 5\\
Encoder MLP Head Layers & 3\\
Encoder Embedding Dim & 128 (corresponding $\sim 11\times11$ environment size)\\
Decoder MLP Layers & 4\\

\end{tabular}
\end{center}
\end{table}

\subsection{Environment Recognition Accuracy During Training}\label{app:EnvAccuracy}
POEM trains more quickly on the environment recognition task and reaches a higher final performance than an equivalent Prototypical Network or Recurrent Network (GRU) (81.1\% vs 72.4\% and 72.1\% ) over a subsequent 100 test episodes.
\begin{figure}[H]
  \centering
  \includegraphics[scale=0.6]{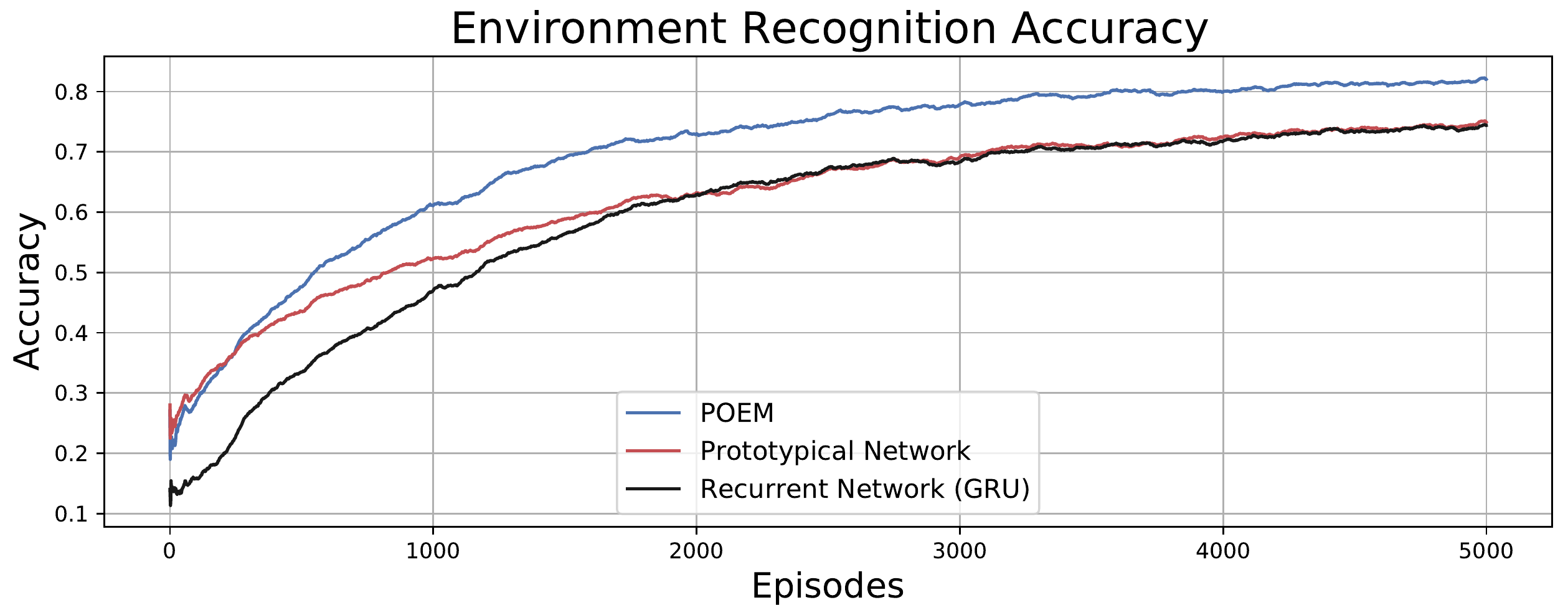}
 
\label{fig:env_classification_accuracy}
\end{figure}

\end{document}